\documentclass[runningheads]{llncs}

\usepackage{eccv}

\usepackage{eccvabbrv}

\usepackage{graphicx}
\usepackage{booktabs}

\usepackage{multirow}

\usepackage[accsupp]{axessibility}  

\usepackage{array}

\usepackage{hyperref}

\usepackage{orcidlink}

\usepackage{pifont}

\newcommand{\cmark}{\ding{51}} 
\newcommand{\xmark}{\ding{55}} 

\begin{document}

\title{Flow Straight to Reality: Perceptually Consistent Flow Matching for Efficient Image Restoration} 

\titlerunning{Perceptually Consistent Flow Matching for Efficient Image Restoration}

\author{Sangwoo Jo\inst{1}\orcidlink{0000-0001-5403-9056} \and Donggeun Ko\inst{2}\orcidlink{0000-0001-5255-4791} \and
Jayeon Kang\inst{1}\orcidlink{0009-0006-5653-0571} \and
Youngsang Kwak\inst{2}\orcidlink{0000-0002-2325-0462} \and
Jaehwa Kwak\inst{2} \and
Sungjoon Choi\inst{1}\orcidlink{0000-0002-9885-333X}}

\authorrunning{S. Jo et al.}

\institute{Korea University, Seoul, South Korea \\
\email{\{jasonjo97, nature0213, sungjoonc\}@korea.ac.kr} \and Aim Future, Seoul, South Korea \\
\email{\{sean.ko, youngsang.kwak, jaehwa.kwak\}@aimfuture.ai}}

\maketitle

\begin{abstract}
Image restoration is fundamentally constrained by the tradeoff between distortion and perception: minimizing pixel-wise error yields over-smoothed results, whereas optimizing for perceptual realism often introduces structural deviations. Recent approaches attempt to balance this tradeoff via posterior sampling or multi-stage generative pipelines, yet remain computationally expensive and architecturally complex. To overcome these limitations, we propose PCFlow (Perceptually Consistent Flow Matching), a unified framework that directly parameterizes a continuous transport from degraded observations to clean targets, jointly optimizing distortion and perceptual quality. While its latent consistency flow objective drives stable and efficient few-step inference, a Latent Consistency Perceptual Loss (LCPL) imposes semantic constraints directly on the guiding velocity field, steering the dynamics toward visually sharp data manifolds. Furthermore, recognizing the inherent conflict between structural and perceptual consistencies, we integrate a conflict-free gradient projection strategy to stabilize the multi-objective optimization landscape. Combined with lightweight, convolution-only backbone, PCFlow achieves competitive performance across diverse restoration tasks at a fraction of traditional computational costs.
\end{abstract}

\section{Introduction}
\label{sec:intro}
Image restoration aims to recover a clean image from its degraded observation, constituting a fundamental class of inverse problems in computer vision. Fundamentally, this process is plagued by the inherent ambiguity of degradation: multiple plausible high-quality reconstructions can map to a single corrupted input. As established in classical literature \cite{blau2018perception, freirich2021theory}, this ambiguity forces a strict \emph{distortion-perception tradeoff}. Methods that minimize distortion error (\eg, MSE) mathematically collapse to the conditional expectation, yielding over-smoothed results. Conversely, approaches optimizing for perceptual realism (\eg, FID) hallucinate high-frequency details, improving visual quality but incurring severe structural deviations and high reconstruction error.

A principled way to achieve perceptually optimal solutions is posterior sampling, which generates samples from the underlying posterior distribution and thus attains the optimal perceptual index \cite{blau2018perception, freirich2021theory}. However, such methods do not minimize distortion and, in expectation, yield an MSE that can be up to twice the minimum achievable error. Recent diffusion and score based restoration approaches leverage generative priors to approximate posterior sampling \cite{saharia2022image, kawar2022denoising, chung2023diffusion, wang2023zeroshot, song2023pseudoinverseguided, zhu2023denoising}. While these approaches significantly improve perceptual quality, they rely on iterative stochastic sampling and typically require multiple sampling steps, resulting in substantial computational cost at inference time. To circumvent this, recent multi-stage frameworks decompose the problem: they first predict a fidelity-oriented minimum mean-squared error (MMSE) estimate, and subsequently apply heavy generative refinement to recover details \cite{ohayon2025posteriormean, Cohen2025BMVC, lin2024diffbir, yue2024difface}. While effective, such two-stage pipelines inherit architectural complexity and still rely on costly generative steps during inference.

\begin{figure*}[t]
    \centering
    \includegraphics[width=0.975\linewidth]{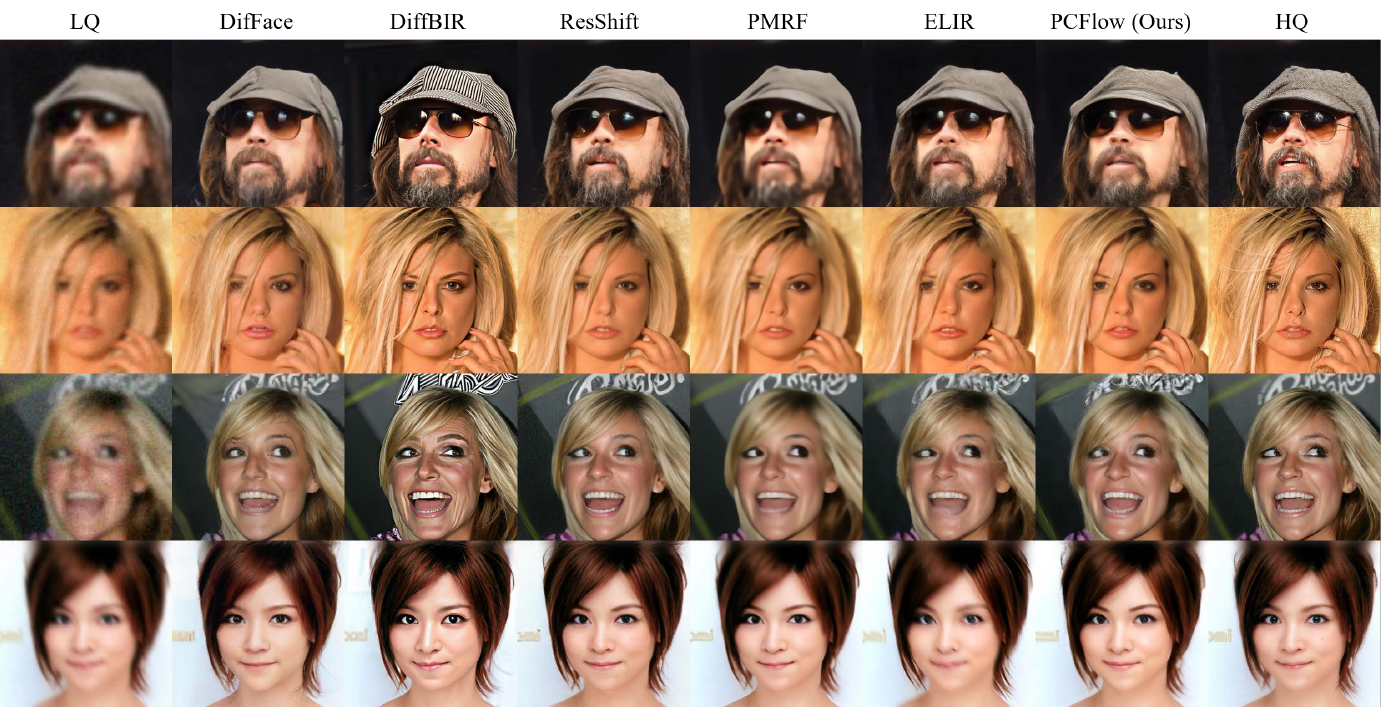}
    \caption{\textbf{Visual Results of PCFlow on the Blind Face Restoration (BFR).} PCFlow consistently produces visually faithful and high-quality reconstructions while requiring only a few inference steps ($K=5$).}
    \label{fig:qualitative_results_bfr}
\end{figure*}

In this work, we revisit the problem from a different perspective. Instead of performing stochastic posterior sampling or decomposing restoration into multiple stages, we propose \textbf{PCFlow} (Perceptually Consistent Flow Matching), a unified direct transport framework that effectively balances distortion and perception. Rather than decomposing the restoration process, we tackle the multi-objective optimization directly within a continuous latent space. 

At its core, PCFlow parameterizes a direct vector field from the degraded input to the clean target. By employing a latent consistency flow matching (LCFM) objective, we enforce geometric consistency along the trajectory, effectively straightening the integration path and enabling image restoration in as few as three inference steps. However, learning a few-step transport with an $L_2$-based objective inherently risks regression to the posterior mean. To mitigate this effect, we introduce a Latent Consistency Perceptual Loss (LCPL), which encourages the trajectory endpoints to align with perceptually sharp, high-density data manifolds. Nevertheless, simply combining these objectives introduces gradient conflicts, particularly at early transport timesteps (low SNR regimes). To stabilize this multi-objective optimization, we adopt a \emph{conflict-free gradient projection} strategy along with an SNR-adaptive schedule, using the perceptual objective as a steering signal while removing structural gradient components that conflict with perceptual optimization.

Furthermore, PCFlow employs a lightweight convolution-only model architecture without attention modules, significantly reducing model size and computational cost. Combined with our conflict-free consistency training, PCFlow operates at a fraction of traditional computational costs while achieving effective balance on the distortion-perception curve.

Our main contributions are summarized as follows:
\begin{itemize}
    \item We propose PCFlow, a unified direct transport framework that integrates perceptual consistency into a latent flow objective, enabling highly efficient few-step image restoration.
    \item We analyze the destructive gradient interference between structural and perceptual objectives, and introduce a conflict-free, SNR-adaptive gradient alignment method to stabilize training.
    \item We adopt a convolution-only architecture that rivals computationally heavy multi-stage diffusion pipelines, significantly reducing parameter count and inference time.
    \item Extensive evaluations across several benchmarks, including blind face restoration, super-resolution, denoising, inpainting, and colorization, demonstrate that PCFlow advances the distortion-perception tradeoff frontier.
\end{itemize}

\section{Related Work}
\subsection{Generative Models for Image Restoration}
Image restoration has undergone a paradigm shift with the advent of generative modeling. Primary works have formulated the problem through the lens of Bayesian posterior sampling \cite{saharia2022image, kawar2022denoising, chung2023diffusion, wang2023zeroshot, song2023pseudoinverseguided, zhu2023denoising}. By iteratively drawing samples from the posterior $p(x\,|\,y) \propto p(x)\, p(y\,|\,x)$ using diffusion priors, these models attain exceptional perceptual quality. Beyond explicit posterior sampling, conditional generative frameworks have been also explored \cite{zhu2024flowie, lin2024diffbir}. Instead of deriving a posterior via likelihood models, these approaches introduce various mechanisms to incorporate degraded images as conditional inputs.

Recent works attempt to bridge distortion minimization and perceptual realism through a disjointed, two-stage pipeline \cite{ohayon2025posteriormean, Cohen2025BMVC}. Motivated by the distortion-perception tradeoff theory \cite{blau2018perception, freirich2021theory}, they first anchor the generation to the minimum mean-squared error (MMSE) estimator,
\begin{equation}
x^* = \mathbb{E}[x \mid y],
\end{equation}
which mathematically guarantees minimal distortion. Subsequently, they must learn an optimal transport from this safe, over-smoothed estimate $x^*$ to the target distribution $x$. To execute this transport, approaches like DOT \cite{adrai2023deep} attempt to bypass iterative sampling by directly approximating the transport trajectory with a closed-form solution under Gaussian assumptions. Although effective, such two-stage pipelines introduce additional architectural complexity and computational cost at inference time.

\subsection{Consistency Flow Matching}
Flow-based generative models \cite{albergo2023building, lipman2023flow, liu2023flow} have recently emerged as an alternative class of generative models that learn a continuous vector field $v(\mathbf{x},t)$ transporting samples between two distributions $\mathbf{x}_0 \sim \pi_0$ and $\mathbf{x}_1 \sim \pi_1$ through an ordinary differential equation (ODE):
\begin{equation}
\frac{d\mathbf{x}}{dt} = v(\mathbf{x}(t),t),
\end{equation}
In other words, flow-based approaches directly parameterize the transport vector field between distributions, enabling  explicit trajectory modeling and efficient inference through numerical ODE integration.

To further improve training and inference efficiency, Consistency Flow Matching (CFM) \cite{yang2025consistency} introduces a consistency objective that effectively straightens the flow trajectory by aligning predictions across neighboring timesteps. Instead of explicitly modeling the entire probability path, CFM enforces consistency in both trajectory outputs and velocity field, leading to stable training and supporting few step inference. While CFM has demonstrated strong performance in unconditional generative modeling, it has primarily been studied in pixel space. Its extension to conditional generative tasks, such as image restoration, as well as to latent representation spaces remains relatively underexplored.

\subsection{Perceptual Objective}
Perceptual objectives have been widely adopted in image restoration to enhance high-frequency details \cite{ledig2017photo, wang2018esrgan, wang2021towards, gu2022vqfr, zhou2022towards}. A common approach is to incorporate feature-based perceptual losses, most notably LPIPS \cite{zhang2018unreasonable}, which measures the distance between pretrained VGGNet features \cite{simonyan2015vgg} extracted from predicted and ground truth images. Such external perceptual losses encourage semantic similarity beyond pixel-wise fidelity and have become standard regularizers in image restoration tasks. E-LatentLPIPS \cite{kang2024distilling} extends LPIPS to latent space, where the training objective remains identical but with additional data augmentation to address the suboptimal loss landscape inherent in latent representations.

More recently, several works have explored leveraging internal features of generative models themselves for perceptual supervision, instead of relying on external pretrained networks. These methods utilize intermediate features, such as U-Net midblock layer or decoder features, to define self-perceptual losses \cite{lin2023diffusion, berrada2025boosting}. By aligning intermediate features, such approaches aim to better preserve generative priors and semantic coherence. 
However, such methods have been developed primarily for image generation, and their extension to restoration settings remains limited.

\section{Method}

\begin{figure*}[t]
    \centering
    \includegraphics[width=1.0\linewidth]{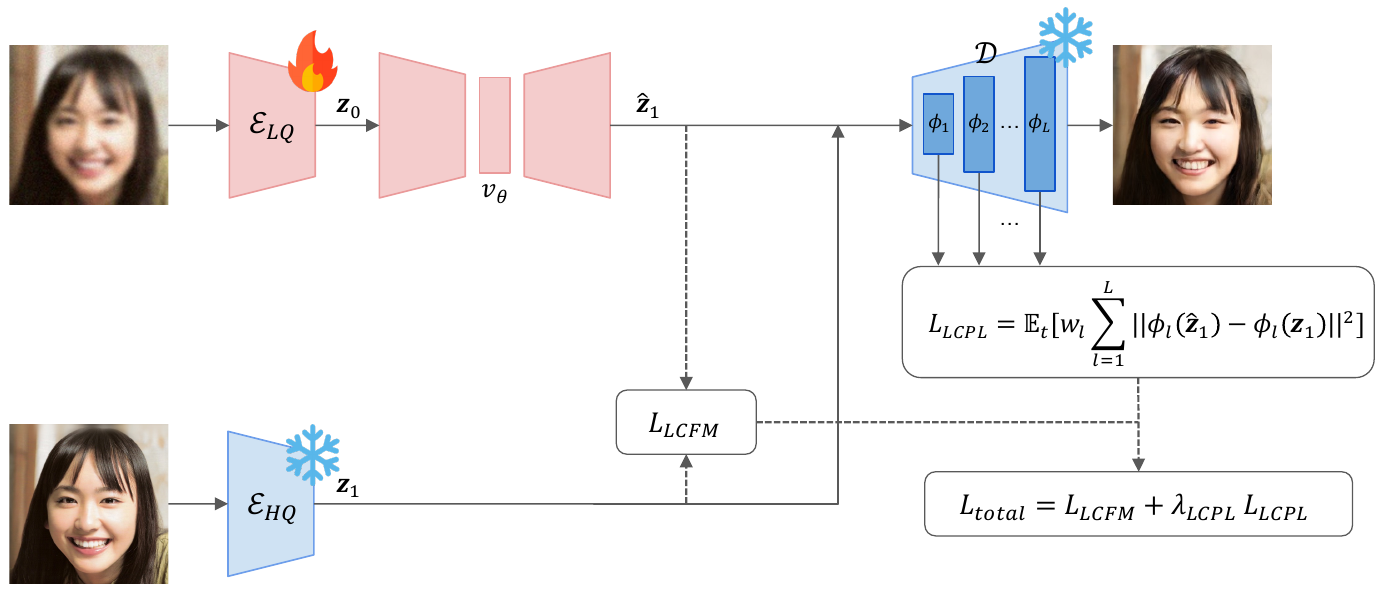}
    \caption{\textbf{Overview of PCFlow.} Given a degraded image, our proposed PCFlow encodes it into latent $\mathbf{z}_0$ using the LQ encoder, and learns the latent transport with a flow model $v_\theta$. The restored latent prediction $\hat{\mathbf{z}}_1$ is then decoded by $\mathcal{D}$ to obtain the final output image $\hat{\mathbf{x}}$. The training objective combines a latent consistency flow matching loss $L_{\mathrm{LCFM}}$ with a latent consistency perceptual loss $L_{\mathrm{LCPL}}$, computed from multi-level decoder features $\{\phi_l\}_{l=1}^{L}$, yielding the final objective as $L_{\mathrm{total}} = L_{\mathrm{LCFM}} + \lambda_{\mathrm{LCPL}}L_{\mathrm{LCPL}}$.} 
    \label{fig:main_fig}
\end{figure*}

\subsection{Latent Consistency Flow Matching}
\medskip
\noindent
\textbf{Formulating Latent Transport.} Following ELIR \cite{Cohen2025BMVC}, we begin by defining a latent consistency flow matching (LCFM) objective that integrates latent flow matching \cite{dao2023flow} with consistency training \cite{yang2025consistency}. Given a low-quality (LQ) latent $\mathbf{z}_0$ and its corresponding high-quality (HQ) latent $\mathbf{z}_1$, we parameterize a vector field $v_{\theta}$ that governs the latent transport over time $t \in [0,1]$:
\begin{equation}
\frac{dz}{dt} = v_{\theta}(\mathbf{z}(t),t),
\end{equation}
where we define a linear interpolation path in the latent space:
\begin{equation}
\mathbf{z}_t = t\mathbf{z}_1 + \left(1 - (1 - \sigma_{\mathrm{min}})\,t\right)\mathbf{z}_0, \quad t \in [0,1].
\end{equation}
Here, $\sigma_{\mathrm{min}} > 0$ prevents degeneration at early timesteps and stabilizes transport learning.

Given the number of segments $K$, we partition the time interval $[0,1]$ into $K$ subintervals $\left\{\left[\frac{i}{K}, \frac{i+1}{K}\right]\right\}_{i=0}^{K-1}$, and let $i = \lfloor Kt \rfloor$ denote the segment index. The LCFM objective then penalizes discrepancies in both the predicted trajectory endpoints and the velocity fields:
\begin{equation}
L_{\text{LCFM}}(\theta) = \mathbb{E}_{\mathbf{z}_0,\mathbf{z}_1, t}\left[\Delta f_{\theta}^{i}(\mathbf{z}_t, \mathbf{z}_{t+\Delta t}, t) + \alpha \Delta v_{\theta}^{i}(\mathbf{z}_t, \mathbf{z}_{t+\Delta t}, t) \right],
\end{equation}
where $t \sim \mathcal{U}(0,1-\Delta t)$ and the penalty terms are defined as:
\begin{align}
\Delta f_{\theta}^{i}(\mathbf{z}_t, \mathbf{z}_{t+\Delta t}, t) &= \left\|f_{\theta}^{i}(\mathbf{z}_t, t) - \text{sg}\left(f_{\theta}^{i}(\mathbf{z}_{t+\Delta t}, t+\Delta t)\right)\right\|^2, \\ 
\Delta v_{\theta}^{i}(\mathbf{z}_t, \mathbf{z}_{t+\Delta t}, t) &= \left\|v_{\theta}^{i}(\mathbf{z}_t, t) - \text{sg}\left(v_{\theta}^{i}(\mathbf{z}_{t+\Delta t}, t+\Delta t)\right)\right\|^2, \\
f_{\theta}^{i}(\mathbf{z}_t, t) &= \mathbf{z}_t + \left(\frac{i+1}{K} - t\right)v_{\theta}^{i}(\mathbf{z}_t, t).
\end{align}
Here, $v_{\theta}^{i}(\mathbf{z}_t, t)$ denotes the predicted vector field within the $i$-th segment, and 
$f_{\theta}^{i}(\mathbf{z}_t, t)$ represents the predicted trajectory output obtained by a single Euler step toward the end of the segment. $\text{sg}(\cdot)$ denotes the stop-gradient operator, and $\alpha, \Delta t$ are set as hyperparameters.

\medskip
\noindent
\textbf{Unconditional and Conditional Training.}
Following PMRF \cite{ohayon2025posteriormean}, we consider both conditional and unconditional flow models for image restoration. For the unconditional flow model, the flow is initialized from the degraded observation, and the transport is defined as:
\begin{equation}
\mathbf{z}_t = t\mathbf{z}_1 + (1-(1-\sigma_{\mathrm{min}})\,t)\mathbf{z}_0^*, \quad \mathbf{z}_0^* = \mathbf{z}_0 + \sigma_s \epsilon,
\end{equation}
where the vector field $v_\theta(z_t, t)$ learns to transport samples from this initialization toward the clean image distribution. Here, $\sigma_s$ controls the standard deviation of the Gaussian noise that alleviates singular mapping between low and high dimensional manifolds.

For comparison, we also consider conditional flow formulation where the vector field learns from the noise distribution conditioned on the degraded input $\mathbf{z}_0$, formulated as the following:
\begin{equation}
\mathbf{z}_t = t\mathbf{z}_1 + (1-(1-\sigma_{\mathrm{min}})\,t)\mathbf{z}_0^*, \quad \mathbf{z}_0^* \sim \mathcal{N}(0,I),
\end{equation}
where the corresponding velocity field $v_\theta(\mathbf{z}_t, t, \mathbf{z}_0)$ learns to transport samples from the noise distribution toward the target clean image conditioned on $\mathbf{z}_0$.

\subsection{Latent Consistency Perceptual Loss}
While LCFM ensures geometric consistency in latent space, such a method does not explicitly enforce perceptual realism. To incorporate semantic alignment, we introduce a Latent Consistency Perceptual Loss (LCPL).

\medskip
\noindent
\textbf{External Perceptual Network.} Given a predicted latent $\hat{\mathbf{z}}_1$ and the ground-truth latent $\mathbf{z}_1$, we adopt E-LatentLPIPS \cite{kang2024distilling} as an external perceptual objective, where the perceptual distance is computed between the corresponding latent representations by applying differentiable augmentation and measuring feature discrepancies in a pretrained network:
\begin{equation}
L_{\text{external}}(\mathbf{z}_1, \hat{\mathbf{z}}_1) = \mathbb{E}_{\mathbf{z}_0,\mathbf{z}_1,t,\mathcal{T}}\left[\left\|f_{\text{VGG}}\left(\mathcal{T}(\mathbf{z}_1)\right) - f_{\text{VGG}}\left(\mathcal{T}(\hat{\mathbf{z}}_1)\right)\right\|^2\right],
\end{equation}
where $f_{\text{VGG}}$ represents a VGG network pretrained on ImageNet \cite{5206848} and BAPPS dataset \cite{zhang2018unreasonable}, and $\mathcal{T}$ denotes random differentiable augmentations implemented in E-LatentLPIPS. We separately train VGGNet for $256 \times 256$ resolution for our experiments, as the publicly available pretrained model from E-LatentLPIPS is trained on $512 \times 512$ images.

\medskip
\noindent
\textbf{Internal Perceptual Network.} 
Although E-LatentLPIPS \cite{kang2024distilling} provides strong perceptual supervision, it relies on an externally pretrained network and may require additional training for dataset-specific tasks. To reduce dependency on external modules, we additionally adopt an internal perceptual objective function defined by the model itself \cite{berrada2025boosting}.

Specifically, let $\{\phi_l(\mathbf{z})\}_{l=1}^{L}$ denote intermediate decoder features extracted from latent feature $z$. The internal perceptual loss is then defined as:
\begin{equation}
L_{\text{internal}}(\mathbf{z}_1, \hat{\mathbf{z}}_1) = \mathbb{E}_{\mathbf{z}_0,\mathbf{z}_1,t}\left[w_l\sum_{l=1}^{L}\left\|\hat{\phi}_{l}(\mathbf{z}_1) - \hat{\phi}_{l}(\hat{\mathbf{z}}_1)\right\|^2\right],
\end{equation}
where $\hat{\phi}_l(\cdot)$ denotes per-channel normalized features and $\{w_l\}_{l=1}^{L}$ are weight values assigned to each feature layer. Note that the formulation is similar to LPL loss \cite{berrada2025boosting} except that binary map masking is omitted for simplicity.

\medskip
\noindent
\textbf{Perceptual Objective for Latent Consistency Training.}
To incorporate perceptual alignment into consistency learning, we define a Latent Consistency Perceptual Loss (LCPL) that enforces perceptual similarity between adjacent trajectory predictions:
\begin{equation}
L_{\text{LCPL}}(\theta) = \mathbb{E}_{\mathbf{z}_0,\mathbf{z}_1,t}\left[L_{\text{percep}}\left(f_{\theta}^{i}(\mathbf{z}_t, t), f_{\theta}^{i}(\mathbf{z}_{t+\Delta t},t+\Delta t)\right)\right],
\end{equation}
where $t \sim \mathcal{U}(0,1-\Delta t)$ and 
$L_{\text{percep}}$ can be instantiated as either $L_{\text{external}}$ or $L_{\text{internal}}$.
This objective encourages perceptual consistency of the predicted trajectory across neighboring timesteps in latent space.

\medskip
\noindent
\textbf{Overall Objective.}
Hence, our final training objective combines flow consistency and perceptual consistency as the following:
\begin{equation}
\label{eq:overall_objective}
L_{\text{total}}(\theta) = L_{\text{LCFM}}(\theta) + \lambda_{\text{LCPL}} \, L_{\text{LCPL}}(\theta),
\end{equation}
where $\lambda_{\text{LCPL}}$ controls the strength of perceptual steering relative to the structural transport objective.

\subsection{Improving Objective with Conflict-Free Gradient Alignment}
\medskip
\noindent
\textbf{Diagnosing Gradient Conflict.}
We compute the gradients of the latent consistency flow matching objective $\nabla_{\theta} L_{\mathrm{LCFM}}$ and perceptual objective $\nabla_{\theta} L_{\mathrm{LCPL}}$ defined in \cref{eq:overall_objective}. 
We observe that the gradients between the structural and the perceptual objective are highly noise-level dependent: the two gradients conflict in low log-SNR regimes but become increasingly aligned in high log-SNR regimes. This indicates that the two objectives induce conflicting descent directions in the parameter space, particularly in early transport stages. However, naively summing the gradients implicitly assumes 

\begin{equation}
\langle \nabla_{\theta} L_{\mathrm{LCFM}}, \nabla_{\theta} L_{\mathrm{LCPL}} \rangle \geq 0,
\end{equation}
which does not hold in practice. When the inner product is negative, the two objectives produce conflicting optimization directions, leading to destructive gradient interference and unstable optimization. This motivates our \emph{conflict-free} gradient update method combined with noise-aware weighting strategy.

\begin{figure}[t]
    \centering
    \begin{minipage}{0.48\linewidth}
        \centering
        \includegraphics[width=\linewidth]{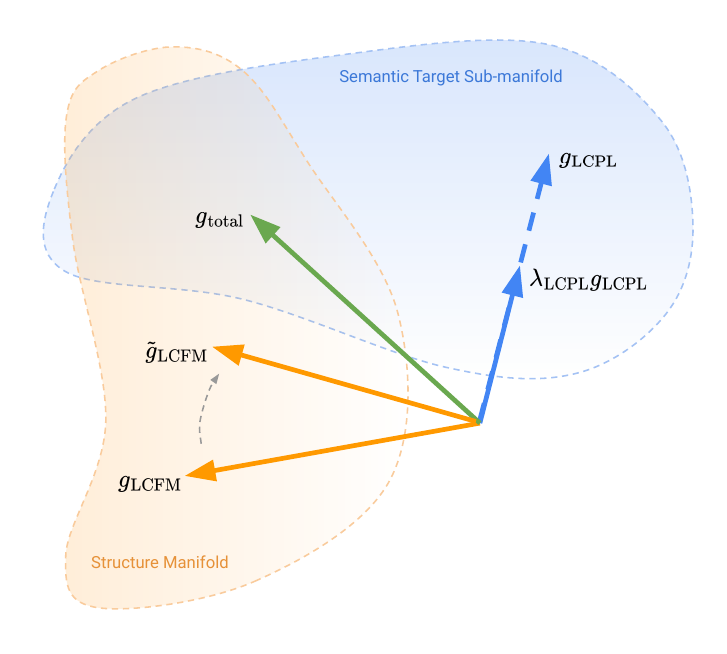}
    \end{minipage}
    \begin{minipage}{0.46\linewidth}
        \centering
        \includegraphics[width=\linewidth]{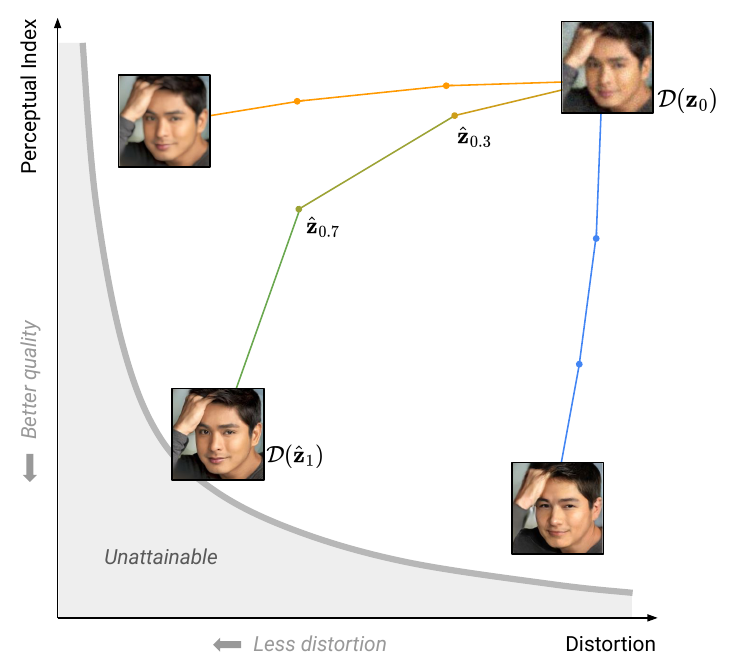}
    \end{minipage}
    \caption{\textbf{Gradient alignment between reconstruction and perceptual objectives.} (Left) The reconstruction objective $L_{\mathrm{LCFM}}$ promotes structural fidelity, while the perceptual objective $L_{\mathrm{LCPL}}$ encourages perceptual realism toward a semantic target sub-manifold. Our gradient alignment method mitigates conflicts between the gradients and produces a balanced update $g_{\mathrm{total}}$. (Right) In the distortion-perception plane, relying solely on LCFM (orange) leads to blurry outputs, whereas over-optimizing LCPL (blue) severely deviates from the input structure and introduces artifacts. In contrast, our aligned trajectory (green) incorporates perceptual guidance while mitigating conflicting structural updates, allowing the model to move closer to the optimal PD limit.}
    \label{fig:main_fig_gradient}
\end{figure}

\medskip
\noindent
\textbf{$\lambda$-scheduling.}
Recognizing that the gradient alignment naturally improves in high SNR regimes (near the clean target), we modulate the perceptual influence dynamically. Instead of a static hyperparameter, we define $\lambda_{\mathrm{LCPL}}(t)$ as a monotonically increasing function of the timestep $t$. Conceptually, this ensures that the model establishes a robust, conflict-free structural foundation during the noisy early stages, and progressively unleashes the full power of perceptual steering as the generation refines toward reality.

\medskip
\noindent
\textbf{Conflict-Free Gradient Update.}
To resolve this destructive interference, we draw inspiration from multi-task gradient surgery \cite{yu2020gradient} but adapt it as an \emph{asymmetric orthogonal projection}. While the LCFM objective provides the structural transport signal, the LCPL objective serves as perceptual guidance that steers the transport trajectory toward perceptually realistic regions of the target manifold. Therefore, when the two vectors conflict (i.e., $\langle g_{\mathrm{LCFM}}, g_{\mathrm{LCPL}} \rangle < 0$), we preserve the perceptual gradient and remove only the component of the structural gradient that conflicts with it.

Denoting the gradients of the two objectives as
$g_{\mathrm{LCFM}} = \nabla_{\theta} L_{\mathrm{LCFM}}$
and
$g_{\mathrm{LCPL}} = \nabla_{\theta} L_{\mathrm{LCPL}}$, we update the parameters as the following procedure:
\begin{equation}
\theta \leftarrow \theta - \eta(\tilde{g}_{\mathrm{LCFM}}(t) + \lambda_{\mathrm{LCPL}}(t)g_{\mathrm{LCPL}}(t)),
\end{equation}
where $\tilde{g}_{\text{LCFM}}$ is defined as the orthogonal component of $g_{\text{LCFM}}$ with respect to $g_{\text{LCPL}}$:
\begin{equation}
\tilde{g}_{\mathrm{LCFM}}(t)
=
g_{\mathrm{LCFM}}(t)
-
\mathbf{1}_{\{\left\langle g_{\mathrm{LCFM}}, g_{\mathrm{LCPL}} \right\rangle<0\}}
\frac{\left\langle g_{\mathrm{LCFM}}, g_{\mathrm{LCPL}} \right\rangle}
{\| g_{\mathrm{LCPL}} \|^2}
\, g_{\mathrm{LCPL}}.
\end{equation}
As a result, the perceptual objective remains an effective steering signal throughout training, whereas reconstruction updates are incorporated only when they do not interfere with perceptual optimization. This asymmetric design enables conflict-free multi-objective optimization and leads to improved perceptual restoration performance. We further provide an ablation study comparing alternative projection strategies in our supplementary material.

\section{Experiments}
We evaluate PCFlow on the following tasks: blind face restoration (BFR), super-resolution, denoising, inpainting, and colorization. Following previous baselines \cite{Cohen2025BMVC, ohayon2025posteriormean}, for BFR, models are trained on FFHQ $512 \times 512$ dataset  and evaluated on CelebA-Test, LFW-Test and CelebAdult. For the remaining tasks, models are trained on FFHQ $256 \times 256$ dataset and evaluated on CelebA-Test dataset.

\subsection{Implementation Details}
\medskip
\noindent
\textbf{Training.} 
We train PCFlow using a two-stage procedure. Initially, the model focuses solely on reconstruction quality by setting $\lambda_{\mathrm{LCPL}} = 0$ for the first 250 training epochs. In the second stage, we enable the perceptual objective and continue training for an additional 250 epochs with the proposed SNR-adaptive weighting, encouraging fine-grained details and perceptual realism. We experiment with different $\lambda$-scheduling strategies and report the best-performing configuration. For consistency training, we follow ELIR setting the number of consistency steps $K=5$ for BFR and $K=3$ for the remaining tasks, with a fixed time interval $\Delta t = 0.05$. We use a batch size of 128 using AdamW optimizer ($\beta_1=0.9$, $\beta_2=0.999$) with weight decay $0.02$. We use exponential moving average (EMA) with decay rate $0.999$ throughout training. 

\medskip
\noindent
\textbf{Model Architecture.} We employ a pretrained Tiny AutoEncoder \cite{von-platen-etal-2022-diffusers}, a light-weight version of Stable Diffusion VAE \cite{esser2024scaling}. The encoder-decoder operates in a 16-channel latent space and contains approximately 2.4M parameters, providing efficient latent representation while maintaining high reconstruction fidelity. We implement convolution-only U-Net architecture from ELIR. The model takes 32 input channels for conditional setting and 16 input channels for unconditional setting. Note that the overall architecture is identical to ELIR, but without the MMSE estimator module, resulting in reduction of 5.5M parameters.

\medskip
\noindent
\textbf{Latent Perceptual Network.} For the internal perceptual network, we extract features from the decoder's intermediate representations, including the mid-block, three upsampling stages, and the final output layer. The contribution of each feature level is weighted according to its spatial resolution, except for the final output layer, \ie, $w_l \propto 2^{-r_l}$, where $r_l$ denotes the resolution level of feature $l$. The weights are subsequently normalized to sum to one. For the external perceptual network, to compute the perceptual objective within our customized latent space, we separately train the model following the procedure from E-LatentLPIPS. In addition, we adopt batch normalization which empirically demonstrates more stable training compared to group normalization.

\begin{table}[t]
\centering
\setlength{\tabcolsep}{3pt}
\renewcommand{\arraystretch}{1.05}
\caption{\textbf{Quantitative results on BFR.} Quantitative comparison of PCFlow with baselines on blind face restoration (BFR) task. The best and second-best results are highlighted in \textbf{bold} and \underline{underlined}, respectively. PCFlow achieves a favorable restoration quality with efficient model architecture and fast inference, obtaining state-of-the-art FID and NIQE scores for CelebA-Test, and FID for CelebAdult dataset. }
\label{tab:quantitative_results_bfr}

\resizebox{\textwidth}{!}{
  \begin{tabular}{lcccccccccc}
  \toprule
  \multirow{3}{*}{Model}
  & \multicolumn{2}{c}{\multirow{2}{*}{Efficiency}}
  & \multicolumn{6}{c}{CelebA-Test} 
  & \multicolumn{1}{c}{LFW} 
  & \multicolumn{1}{c}{CelebAdult} \\
  \cmidrule(lr){4-9}\cmidrule(lr){10-10}\cmidrule(lr){11-11}
  & \multicolumn{2}{c}{}
  & \multicolumn{3}{c}{Perceptual Quality}
  & \multicolumn{3}{c}{Distortion} 
  & \multicolumn{2}{c}{Perceptual Quality} \\
  \cmidrule(lr){2-3}\cmidrule(lr){4-6} \cmidrule(lr){7-9} \cmidrule(lr){10-10} \cmidrule(lr){11-11}
  & \#Params[M]$\downarrow$ & FPS$\uparrow$ 
  & FID$\downarrow$ & NIQE$\downarrow$ & MUSIQ$\uparrow$ & PSNR$\uparrow$ & SSIM$\uparrow$ & LPIPS$\downarrow$ & FID$\downarrow$ & FID$\downarrow$ \\
  \midrule

  CodeFormer & 94 & 27.13 & 52.23 & 4.65 & \textbf{75.55} & 24.77 & 0.6732 & \textbf{0.3432} & 54.28 & 114.34 \\
  GFPGAN(v1.3) & 87.14 & \textbf{59.73} & 45.95 & 4.42 & \underline{75.29} & 24.60 & 0.6802 & 0.3643 & 49.58 & 112.31 \\
  VQFRv2 & 83.49 & 16.97 & 46.01 & 4.17 & 74.40 & 22.85 & 0.6446 & 0.3624 & 52.50 & 106.83 \\
  Difface $(K=100)$ & 182.07 & 0.78 & 37.43 & 4.37 & 68.31 & 24.44 & 0.6579 & 0.4172 & \underline{47.23} & \underline{101.11} \\
  DiffBIR $(K=50)$ & 1666.93 & 0.38 & 46.23 & 4.21 & 75.27 & 23.27 & 0.6379 & 0.3876 & \textbf{46.03} & 111.84 \\
  ResShift $(K=4)$ & 196.70 & 6.85 & 43.60 & 4.37 & 72.16 & 25.32 & 0.6965 & \underline{0.3435} & 54.23 & 109.04 \\
  PMRF $(K=25)$ & 182.75 & 0.57 & \underline{37.22} & \underline{4.12} & 70.36 & \textbf{25.85} & \textbf{0.7098} & 0.3470 & 49.98 & 104.44 \\
  ELIR $(K=5)$ & \underline{37.52} & 33.11 & 44.64 & 5.26 & 67.24 & \underline{25.56} & \underline{0.7030} & 0.3735 & 53.19 & 105.55 \\
  \textbf{PCFlow (Ours)} & \textbf{32.02} & \underline{42.62} & \textbf{35.89} & \textbf{3.95} & 70.35 & 24.44 & 0.6680 & 0.3850 & 50.68 & \textbf{98.85} \\
  \bottomrule
  \end{tabular}
  }
\end{table}

\begin{table}[htbp]
\centering
\setlength{\tabcolsep}{3pt}
\renewcommand{\arraystretch}{1.05}
\caption{\textbf{Quantitative results on the remaining restoration tasks.} Quantitative comparison of PCFlow with baselines ELIR and PMRF. The best and second-best results are highlighted in \textbf{bold} and \underline{underlined}, respectively. PCFlow consistently outperforms ELIR in FID, despite using only 21M parameters compared to ELIR (27M) and PMRF (176M), demonstrating favorable perceptual quality with efficient model architecture and fast inference.}
\label{tab:quantitative_results_ir}

\resizebox{\textwidth}{!}{
  \begin{tabular}{llccccccc}
  \toprule
  \multirow{2}{*}{Task} & \multirow{2}{*}{Model}
  & \multicolumn{1}{c}{Efficiency}
  & \multicolumn{1}{c}{Perceptual Quality}
  & \multicolumn{3}{c}{Distortion} \\
  \cmidrule(lr){3-3}\cmidrule(lr){4-4}\cmidrule(lr){5-7}
  & & \#Params[M]$\downarrow$ & FID$\downarrow$ & PSNR$\uparrow$ & SSIM$\uparrow$ & LPIPS$\downarrow$ \\
  \midrule

  \multirow{3}{*}{Super Resolution}
  & PMRF $(K=25)$ & 176 & \textbf{44.64} & \textbf{24.40} & \textbf{0.6708} & \textbf{0.2991} \\
  & ELIR & 27 & 49.25 & \underline{23.57} & 0.6439 & \underline{0.3299} \\
  & \textbf{PCFlow (Ours)} & 21 & \underline{45.50} & 23.38 & \underline{0.6512} & 0.3328 \\
  \midrule

  \multirow{3}{*}{Denoising}
  & PMRF $(K=25)$ & 176 & \textbf{44.35} & \textbf{27.92} & \textbf{0.7782} & \textbf{0.2401} \\
  & ELIR & 27 & 47.70 & \underline{26.67} & \underline{0.7521} & \underline{0.2619} \\
  & \textbf{PCFlow (Ours)} & 21 & \underline{45.42} & 26.26 & 0.7480 & 0.2800 \\
  \midrule

  \multirow{3}{*}{Inpainting}
  & PMRF $(K=25)$ & 176 & \textbf{42.88} & \textbf{26.19} & \textbf{0.7383} & \textbf{0.2626} \\
  & ELIR & 27 & 47.82 & 24.85 & 0.7105 & \underline{0.2840} \\
  & \textbf{PCFlow (Ours)} & 21 & \underline{45.50} & \underline{24.92} & \underline{0.7195} & 0.2936 \\
  \midrule

  \multirow{3}{*}{Colorization}
  & PMRF $(K=25)$ & 176 & \textbf{42.61} & \textbf{23.47} & \underline{0.7122} & \textbf{0.3463} \\
  & ELIR & 27 & 51.72 & \underline{23.15} & 0.7113 & \underline{0.3587} \\
  & \textbf{PCFlow (Ours)} & 21 & \underline{45.21} & 22.18 & \textbf{0.7499} & 0.3596 \\
  \bottomrule
  \end{tabular}
  }
\end{table}

\subsection{Quantitative Results}
\medskip
\noindent
\textbf{Blind Face Restoration.} Quantitative results on blind face restoration (BFR) are reported in \cref{tab:quantitative_results_bfr}. PCFlow achieves state-of-the-art perceptual quality, obtaining the best FID and NIQE on CelebA-Test and the best FID on CelebAdult, while requiring only 32M parameters and five sampling steps $(K=5)$. Compared to ELIR, PCFlow uses fewer parameters (32M vs. 37.5M) and achieves 1.29$\times$ higher inference speed (42.62 vs. 33.11 FPS), while substantially improving FID (35.89 vs. 44.64) on CelebA-Test. Furthermore, despite being significantly smaller and faster than diffusion-based baselines such as PMRF (183M parameters, 0.57 FPS), PCFlow outperforms in perceptual quality metrics while delivering over 75$\times$ higher throughput. These results demonstrate that PCFlow provides a favorable trade-off between restoration quality and computational efficiency, achieving strong perceptual restoration performance under a practical few-step inference regime.

\medskip
\noindent
\textbf{Other Restoration Tasks.} We further provide quantitative results for the remaining restoration tasks, including super-resolution, denoising, inpainting, and colorization, in \cref{tab:quantitative_results_ir}. PCFlow consistently improves over ELIR in FID across all four image restoration tasks. These results challenge the prevailing paradigm that prioritizes explicit posterior mean estimation followed by transport refinement. Instead, we show that directly learning the conditional transport from degraded observations, combined with perceptual supervision as a steering signal, is sufficient to achieve superior distortion-perception trade-offs under a few-step regime. Notably, the improvements are consistent across diverse degradation types, suggesting that the proposed formulation generalizes well beyond a specific restoration setting.

\begin{figure}[t]
    \centering
    \includegraphics[width=\linewidth]{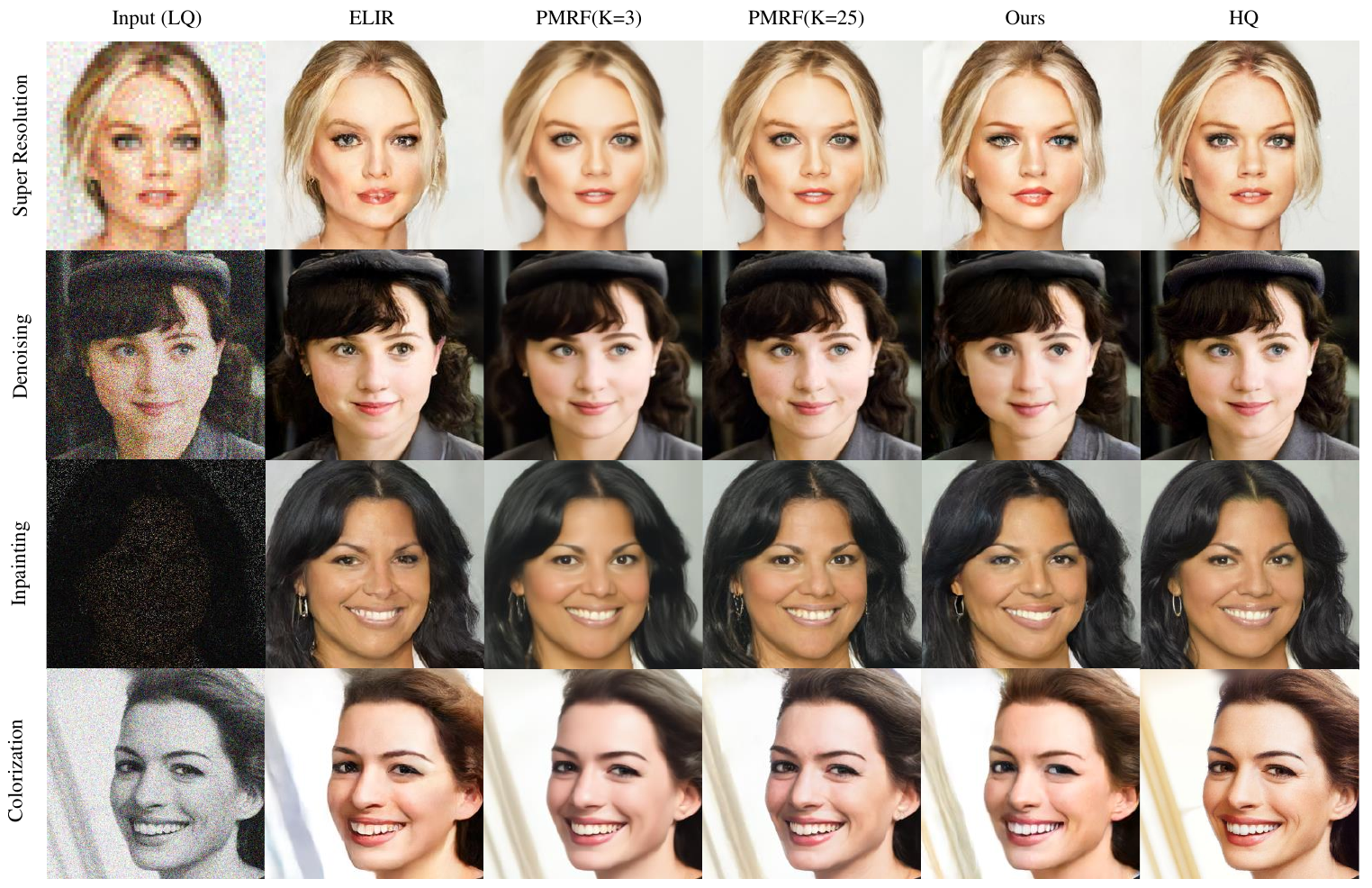}
    \caption{\textbf{Qualitative comparison across image restoration tasks.} From left to right: Input (LQ), ELIR, PMRF$(K=3)$, PMRF$(K=25)$, PCFlow (Ours), and ground truth image (HQ). PCFlow produces visually sharp and coherent reconstructions even with a small number of integration steps $K=3$, establishing an effective training framework that provides a favorable distortion-perception trade-off.}
    \label{fig:qualitative_results}
\end{figure}

\subsection{Qualitative Results}
\medskip
\noindent
\textbf{Blind Face Restoration.}
\cref{fig:qualitative_results_bfr} presents qualitative comparisons on blind face restoration (BFR). Compared with ELIR, PCFlow consistently restores sharper facial structures, including hair, beard, and facial contours, while maintaining more natural textures and local contrast. These improvements lead to visually more faithful reconstructions without introducing noticeable artifacts. Compared with diffusion-based methods, PCFlow produces visually balanced reconstructions without the tendency toward over-sharpened textures observed in DiffBIR, while achieving visual quality competitive with DifFace and PMRF. Notably, these results are obtained using only five inference steps, demonstrating that the proposed framework effectively achieves high perceptual restoration quality with highly efficient inference.

\medskip
\noindent
\textbf{Other Restoration Tasks.}
\cref{fig:qualitative_results} presents qualitative comparisons across four image restoration tasks. Compared to ELIR, PCFlow consistently generates visually more coherent and realistic reconstructions. In particular, ELIR tends to generate slightly over-smoothed facial structures, while PCFlow restores sharper facial boundaries and more consistent skin textures, leading to perceptually more faithful reconstructions. Note that PCFlow can occasionally generate subtle artifacts around high-frequency regions such as the eyes. Compared to PMRF with a larger number of integration steps $K=25$, PCFlow attains strong perceptual quality while operating in a significantly more efficient few-step regime. These observations align with the quantitative trends in \cref{tab:quantitative_results_ir}, where PCFlow improves perceptual metrics over ELIR and achieves a favorable distortion-perception trade-off, despite requiring substantially fewer transport steps and model parameters than PMRF$(K=25)$.

\subsection{Ablation Studies}
We conduct several ablation studies and analyze the contribution of each component in our proposed PCFlow training objective.

\medskip
\noindent
\textbf{Preheating Period.}
\cref{tab:preheating_period} evaluates the contributions of the proposed preheating period. The preheating period corresponds to the first 250 epochs, during which only the reconstruction objective is optimized while perceptual supervision is disabled. Jointly training both distortion and perception objectives from the beginning yields inferior results compared to PCFlow, suggesting that the initial preheating stage is essential for establishing a stable coarse-to-fine transport trajectory before introducing perceptual supervision. In addition, applying conflict-free gradient alignment from epoch 0 consistently improves FID over naive joint optimization (46.40 $\rightarrow$ 46.26), indicating that mitigating optimization conflicts between reconstruction and perceptual objectives is beneficial. Combining both components yields the best performance across all metrics.

\begin{table}[t]
\centering
\caption{\textbf{Validation of the Preheating Period.} The combination of preheating period, gradient surgery, and linear warmup schedule demonstrates the best overall performance in both distortion and perceptual metrics on super-resolution task.}
\label{tab:preheating_period}

\resizebox{\columnwidth}{!}{
    \begin{tabular}{lccc>{\centering\arraybackslash}p{0.9cm}cccc}
    \toprule
    Setting & $L_{\mathrm{LCPL}}$ & Preheating & $\lambda$-scheduling & FID$\downarrow$ & PSNR$\uparrow$ & SSIM$\uparrow$ & LPIPS$\downarrow$ \\
    \midrule
    $L_{\mathrm{LCFM}}$ only & -- & -- & -- & 46.10 & 23.35 & 0.6476 & 0.3373 \\
    w/o Gradient Alignment & epoch 0 & \texttimes & linear warmup & 46.40 & 23.32 & 0.6484 & 0.3363 \\
    w/ Gradient Alignment & epoch 0 & \texttimes & linear warmup & 46.26 & 23.31 & 0.6480 & 0.3364 \\
    PCFlow & epoch 250 & \checkmark & linear warmup & \textbf{45.50} & \textbf{23.38} & \textbf{0.6512} & \textbf{0.3328} \\
    \bottomrule
    \end{tabular}
}
\end{table}

\begin{table}[t]
\centering
\setlength{\tabcolsep}{3pt}
\caption{\textbf{Ablation on PCFlow components.} Quantitative results for adding each PCFlow component on colorization task. The best performance under conditional flow formulation (B--E) is highlighted in \textbf{bold}, while the second-best are \underline{underlined}.}
\label{tab:ablation_pcflow_components}

\resizebox{0.9\textwidth}{!}{
\begin{tabular}{lcccc}
    \toprule
    & FID$\downarrow$ 
    & PSNR$\uparrow$
    & SSIM$\uparrow$ 
    & LPIPS$\downarrow$ \\
    \midrule
    
    Base (A)
    & 56.27 & 23.11 & 0.7651 & 0.3675 \\
    
    + Conditional (B)
    & 47.21 & 21.88 & 0.7220 & 0.3845 \\
    
    + Encoder Fine-Tuning (C)
    & 45.97 & \underline{22.19} & 0.7465 & 0.3654 \\
    
    + Perceptual Objective $(L_{\mathrm{LCPL}})$ (D)
    & \underline{45.78} & \textbf{22.27} & \textbf{0.7530} & \textbf{0.3551} \\
    
    + Conflict-Free Gradient Alignment (E)
    & \textbf{45.21} & 22.18 & \underline{0.7499} & \underline{0.3596} \\
    \bottomrule
\end{tabular}
}
\end{table}

\medskip
\noindent
\textbf{Individual Training Components.}
\cref{tab:ablation_pcflow_components} shows quantitative results for progressively adding each of the PCFlow components. For the baseline unconditional model (config A), we set $\sigma_{\mathrm{min}}$ to 0.05 for super-resolution, 0.01 for denoising and colorization, and 0.025 for inpainting task. Training with conditional flow matching formulation (config B) improves FID by a substantial margin from 56.27 to 47.21. Fine-tuning the encoder along with the vector field (config C) further enhances model performance in both distortion and perceptual metrics. Adding the perceptual objective (config D) improves perceptual quality while maintaining reconstruction fidelity, effectively steering the model towards perceptually sharp data manifolds. Finally, incorporating the proposed conflict-free gradient alignment (config E) achieves the lowest FID of 45.21, pushing the model closer to the optimal distortion--perception tradeoff. We observe that similar trends consistently hold across the remaining image restoration tasks. 

\medskip
\noindent
\textbf{$\lambda$-scheduling.}
We further investigate the role of the perceptual weight $\lambda_{\mathrm{LCPL}}$ as a function of the diffusion timestep $t$. \cref{tab:ablation_lambda_scheduling} shows quantitative comparison of different scheduling methods, including constant, linear, and linear warmup schedules. We observe that linear warmup schedule consistently achieves the best FID across all four image restoration tasks, demonstrating the benefit of gradually introducing perceptual supervision during transport. We hypothesize that early transport steps primarily require stable global alignment between degraded observations and the target distribution, where excessive perceptual supervision may introduce unnecessary optimization conflicts. In contrast, as the transport progresses closer to the true image manifold, perceptual supervision becomes more beneficial for recovering fine structures and visually realistic details. 

\begin{table*}[t]
\centering
\footnotesize
\setlength{\tabcolsep}{4pt}
\renewcommand{\arraystretch}{1.12}
\caption{\textbf{Ablation on $\lambda$-scheduling.} Comparison of different scheduling methods for $\lambda_{\mathrm{LCPL}}$. Increasing the perceptual weight via linear warmup schedule consistently achieves the best model performance in FID across four image restoration tasks.}
\label{tab:ablation_lambda_scheduling}

\begin{tabular}{l l cc}
    \toprule
    Task & $\lambda$-scheduling
    & FID$\downarrow$ 
    & LPIPS$\downarrow$ \\
    \midrule
    
    \multirow{3}{*}{Super-Resolution}
    & constant & 46.18 & \textbf{0.3300} \\
    & linear & \underline{45.74} & \underline{0.3326} \\
    & linear warmup & \textbf{45.50} & 0.3328 \\
    \midrule
    
    \multirow{3}{*}{Denoising}
    & constant & 46.12 & \textbf{0.2790} \\
    & linear & \underline{45.48} & 0.2814 \\
    & linear warmup & \textbf{45.42} & \underline{0.2800} \\
    \midrule
    
    \multirow{3}{*}{Inpainting}
    & constant & 46.25 & \textbf{0.2918} \\
    & linear & \underline{45.56} & 0.2944 \\
    & linear warmup & \textbf{45.50} & \underline{0.2936} \\
    \midrule
    
    \multirow{3}{*}{Colorization}
    & constant & 45.78 & \textbf{0.3551} \\
    & linear & \underline{45.45} & 0.3611 \\
    & linear warmup & \textbf{45.21} & \underline{0.3596} \\
    \bottomrule
\end{tabular}
\end{table*}

\medskip
\noindent
\subsection{Analysis on Gradient Conflict}
\begin{figure*}[t]
    \centering
    \includegraphics[width=0.92\linewidth]{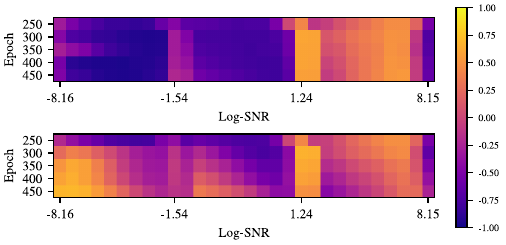}
    \caption{\textbf{Analysis of gradient interaction between flow matching and perceptual objectives.} Cosine similarity maps between the LCFM and LCPL gradients as a function of training epochs and log-SNR, shown without (top) and with (bottom) the proposed conflict-free update. Our proposed method significantly mitigates gradient conflict, particularly in low-SNR regions, and promotes consistent optimization throughout the sampling process.}
    \label{fig:ablation_grad_conflict}
\end{figure*}

We analyze the interaction between reconstruction objective $\mathcal{L}_{\mathrm{LCFM}}$ and perceptual objective $\mathcal{L}_{\mathrm{LCPL}}$ by comparing the results from standard joint optimization with our proposed gradient alignment method. As illustrated in \cref{fig:ablation_grad_conflict}, the cosine similarity heatmap reveals substantial regions of negative alignment (blue) under the standard update, indicating that the two objectives frequently produce conflicting gradients. Such conflicts can hinder stable optimization and slow down convergence. In contrast, after applying the proposed conflict-free update, the gradients exhibit significantly improved alignment, with fewer negatively correlated regions and more consistent positive interactions.

\section{Conclusion}
In this paper, we propose PCFlow, a unified latent transport framework for image restoration that jointly optimizes distortion and perceptual quality. By formulating the problem with latent consistency flow matching objective, our method directly learns the transport between degraded and clean images, hence enabling efficient few-step inference. To enhance perceptual realism, we additionally introduce a latent consistency perceptual objective that enforces semantic alignment along the transport trajectory. Furthermore, motivated by our analysis of the interaction between flow consistency and perceptual objectives, we propose a conflict-free gradient update method that enables the perceptual objective to steer optimization while removing conflicting structural gradient components. Extensive experiments demonstrate that PCFlow establishes a more favorable distortion--perception trade-off frontier compared to previous baselines across various image restoration tasks, while requiring only a few inference steps and maintaining an efficient model architecture.


\section*{Acknowledgements}
This work was partly supported by the Institute of Information $\&$ Communications Technology Planning $\&$ Evaluation (IITP)-ITRC (Information Technology Research Center) grant funded by the Korea government (MSIT) (IITP-2026-RS-2024-00436857, 50$\%$) and by IITP grant funded by the Korea government (MSIT) (No. RS-2019-II190079, Artificial Intelligence Graduate School Program (Korea University), 50$\%$).

%
%
\bibliographystyle{splncs04}
\bibliography{main}

\clearpage
\appendix






\title{Supplementary Material for \\ Flow Straight to Reality: Perceptually Consistent Flow Matching for Efficient Image Restoration}

\titlerunning{Perceptually Consistent Flow Matching for Efficient Image Restoration}

\author{Sangwoo Jo\inst{1}\orcidlink{0000-0001-5403-9056} \and Donggeun Ko\inst{2}\orcidlink{0000-0001-5255-4791} \and
Jayeon Kang\inst{1}\orcidlink{0009-0006-5653-0571} \and
Youngsang Kwak\inst{2}\orcidlink{0000-0002-2325-0462} \and
Jaehwa Kwak\inst{2} \and
Sungjoon Choi\inst{1}\orcidlink{0000-0002-9885-333X}}

\authorrunning{S. Jo et al.}

\institute{Korea University, Seoul, South Korea \\
\email{\{jasonjo97, nature0213, sungjoonc\}@korea.ac.kr} \and Aim Future, Seoul, South Korea \\
\email{\{sean.ko, youngsang.kwak, jaehwa.kwak\}@aimfuture.ai}}

\maketitle


\section{Additional Implementation Details}
\subsection{Hyperparameters}
We provide training hyperparameters for our proposed PCFlow in \cref{tab:hyperparams}. Note that although PCFlow can be trained within a single unified procedure, we separate the stages in practice to allow additional flexibility for hyperparameter tuning. For the convolution-only U-Net backbone, we set the number of channels for the downsampling blocks as $[128,256,256,512]$ and number of mid blocks to 3 for blind face restoration (BFR) and 1 for the remaining tasks, following ELIR \cite{Cohen2025BMVC}. For the LCPL objective, we use decoder features as a source of perceptual supervision, similar to the LPL loss formulation \cite{berrada2025boosting}. We set the perceptual objective coefficient $\lambda_{\mathrm{LCPL}}$ to increase over timestep $t$, \ie,
\begin{equation}
\lambda_{\mathrm{LCPL}}(t)=\lambda_{\min}+\mathbb{I}_{t \geq t_{\min}}(\lambda_{\max}-\lambda_{\min})\left(\frac{t-t_{\min}}{1-t_{\min}}\right),
\end{equation}
which progressively shifts the trajectory towards the target manifold. We set $\lambda_{\min}=0$, $\lambda_{\max}=0.5$, and $t_{\min}=0.5$ for linear warmup schedule.

\subsection{Training Details for E-LatentLPIPS}
In this section, we provide training details for implementing E-LatentLPIPS \cite{kang2024distilling} which serves as an external perceptual supervision in our ablation study. We first train a VGG network \cite{simonyan2015vgg} on ImageNet dataset \cite{5206848} with resolution of~$256 \times 256$ to learn general feature representations in the latent space. We evaluate two variants of the VGG16 backbone with group normalization and batch normalization. We empirically observe that the batch normalization yields higher accuracy, and hence select it as our backbone for the external perceptual network. We subsequently fine-tune the network on BAPPS dataset \cite{zhang2018unreasonable} to align the features with human perceptual similarity judgments. Overall, as illustrated in \cref{tab:elatentlpips}, our implementation achieves similar performance to the original work \cite{kang2024distilling}, despite being trained at resolution of $256 \times 256$ and latent space of the Tiny AutoEncoder \cite{von-platen-etal-2022-diffusers}, which is a compressed version of Stable Diffusion VAE \cite{esser2024scaling}.  

\begin{table}[!hbtp]
\centering
\setlength{\tabcolsep}{3pt}
\caption{Training hyperparameters for the two training stages of PCFlow. The first column corresponds to blind face restoration tasks (BFR), while the second column corresponds to the remaining four restoration tasks, including super-resolution, denoising, inpainting, and colorization.}
\label{tab:hyperparams}

{(a) PCFlow with LCFM objective only.}

\vspace{5pt}
\resizebox{0.8\textwidth}{!}{
\begin{tabular}{lcc}
    \toprule
    Hyperparameters & Blind Face Restoration & Other Restoration tasks \\
    \midrule
    Parameters & 32M & 21M \\
    Euler steps ($M$) & 5 & 3 \\
    CFM segments ($K$) & 5 & 3 \\
    CFM $\Delta t$ & 0.05 & 0.05 \\
    CFM $\alpha$ & 0.001 & 0.001 \\
    $\sigma_\mathrm{min}$ & $10^{-5}$ & $10^{-5}$ \\
    Training epochs & 250 & 250 \\
    Batch size & 32 & 128 \\
    Image dimension & $3 \times 512 \times 512$ & $3 \times 256 \times 256$ \\
    Latent dimension & $16 \times 64 \times 64$ & $16 \times 32 \times 32$ \\
    Precision & bfloat16 mixed & bfloat16 mixed \\
    Training hardware & $1 \times$ H100 80GB & $1 \times$ A100 80GB \\
    Training time & 2 days & 1 day \\
    Optimizer & AdamW & AdamW \\
    Learning rate & $2 \times 10^{-4}$ & $2 \times 10^{-4}$ \\
    AdamW betas & (0.9, 0.999) & (0.9, 0.999) \\
    Weight decay & 0.02 & 0.02 \\
    EMA decay & 0.999 & 0.999 \\
    \bottomrule
\end{tabular}
}

\vspace{10pt}
{(b) PCFlow with perceptual objective (LCPL).}

\vspace{5pt}
\resizebox{0.8\textwidth}{!}{
\begin{tabular}{lcc}
    \toprule
    Hyperparameters & Blind Face Restoration & Other Restoration tasks \\
    \midrule
    Parameters & 32M & 21M \\
    Euler steps ($M$) & 5 & 3 \\
    CFM segments ($K$) & 5 & 3 \\
    CFM $\Delta t$ & 0.05 & 0.05 \\
    CFM $\alpha$ & 0.001 & 0.001 \\
    $\sigma_\mathrm{min}$ & $10^{-5}$ & $10^{-5}$ \\
    Training epochs & 250 & 250 \\
    Batch size & 32 & 128 \\
    Image dimension & $3 \times 512 \times 512$ & $3 \times 256 \times 256$ \\
    Latent dimension & $16 \times 64 \times 64$ & $16 \times 32 \times 32$ \\
    Precision & bfloat16 mixed & bfloat16 mixed \\
    Training hardware & $1 \times$ H100 80GB & $1 \times$ A100 80GB \\
    Training time & 2.5 days & 2 days \\
    Optimizer & AdamW & AdamW \\
    Learning rate & $2 \times 10^{-4}$ & $2 \times 10^{-4}$ \\
    AdamW betas & (0.9, 0.999) & (0.9, 0.999) \\
    Weight decay & 0.02 & 0.02 \\
    EMA decay & 0.999 & 0.999 \\
    $w_l$ & $[1, 0.5, 0.25, 0.125, 1]$ & $[1, 0.5, 0.25, 0.125, 1]$ \\
    $\lambda_{\mathrm{LCPL}}(t)$ & linear warmup & linear warmup \\
    $\lambda_{\min}$ & 0 & 0 \\
    $\lambda_{\max}$ & 0.5 & 0.5 \\
    $t_{\min}$ & 0.5 & 0.5 \\
    \bottomrule
\end{tabular}
}
\end{table}

\begin{table}[htbp]
\centering
\caption{\textbf{ImageNet and BAPPS Classification Scores.} Classification accuracy on ImageNet and BAPPS datasets within the Tiny AutoEncoder latent space. Our reimplementation achieves similar performance to the original E-LatentLPIPS paper.}
\label{tab:elatentlpips}

\setlength{\heavyrulewidth}{0.08em}
\setlength{\tabcolsep}{18pt}

\begin{tabular}{@{}llcc}
\toprule
Dataset & Type & Ours & E-LatentLPIPS (Original) \\
\midrule

ImageNet & VGG-BN & 67.82 & 68.26 \\

\midrule

 & Traditional & 76.62 & 74.29 \\
BAPPS & CNN & 82.43 & 81.99 \\
 & Real & 63.79 & 63.21 \\

\bottomrule
\end{tabular}
\end{table}

\section{Additional Results}
\subsection{Further Ablation Studies}
\medskip
\noindent
\textbf{Unconditional vs. Conditional Flow Models.}
We consider both unconditional and conditional flow formulations. We empirically find that the conditional flow model consistently achieves better FID across all restoration tasks, as shown in \cref{tab:ablation_cond}. Unlike PMRF, which reports limited gains from conditional transport, we observe that initializing from random noise allows the model to fully leverage its generative capability. From an optimal transport perspective, the unconditional formulation may restrict the transport trajectory to remain close to the degraded manifold, whereas the conditional formulation enables transport with generative prior resulting in higher perceptual realism.

\begin{table*}[htbp]
\centering
\footnotesize
\setlength{\tabcolsep}{4pt}
\renewcommand{\arraystretch}{1.12}
\caption{\textbf{Ablation on Unconditional vs. Conditional Flow Models.} Learning a transport from random noise with degraded image as a conditional input leads to substantial improvements in FID over unconditional formulation.}
\label{tab:ablation_cond}

\begin{tabular}{c c cccc}
\toprule
Task & Conditional 
& FID$\downarrow$ 
& LPIPS$\downarrow$ \\
\midrule

\multirow{2}{*}{Super-Resolution}
& \xmark & 63.69 & 0.3509 \\
& \cmark & \textbf{46.42} & \textbf{0.3396} \\
\midrule

\multirow{2}{*}{Denoising}
& \xmark & 48.50 & \textbf{0.2866} \\
& \cmark & \textbf{45.70} & 0.3025 \\
\midrule

\multirow{2}{*}{Inpainting}
& \xmark & 60.47 & 0.3677 \\
& \cmark & \textbf{47.04} & \textbf{0.3611} \\
\midrule

\multirow{2}{*}{Colorization}
& \xmark & 56.27 & \textbf{0.3675} \\
& \cmark & \textbf{47.21} & 0.3845 \\
\bottomrule
\end{tabular}
\end{table*}

\medskip
\noindent
\textbf{Encoder Fine-Tuning.}
We compare freezing the pretrained encoder against jointly fine-tuning it together with the vector field network. As shown in \cref{tab:ablation_ft}, fine-tuning the encoder consistently improves both reconstruction and perceptual quality across all image restoration tasks. For instance, for super-resolution, FID and LPIPS decrease from $46.42$ to $46.10$ and from $0.3396$ to $0.3373$, respectively, while PSNR and SSIM increase from $23.30$ to $23.35$ and from $0.6462$ to $0.6476$, respectively. Similar improvements are observed in other image restoration tasks. We hypothesize that, since the encoder is pretrained on high-quality images, it requires additional training to fully learn the representation of degraded inputs. Joint optimization therefore allows the encoder to adapt to the degradation distribution, resulting in improved alignment between the latent representation and the restoration dynamics.

\begin{table*}[t]
\centering
\footnotesize
\setlength{\tabcolsep}{4pt}
\renewcommand{\arraystretch}{1.12}
\caption{\textbf{Ablation on Encoder Fine-Tuning (FT).} Comparison of PCFlow and with and without encoder fine-tuning. Fine-tuning encoder along with the vector field shows better performance in both distortion and perception metrics across all four image restoration tasks.}
\label{tab:ablation_ft}

\begin{tabular}{l c cccc}
    \toprule
    Task & Encoder FT 
    & FID$\downarrow$ 
    & PSNR$\uparrow$
    & SSIM$\uparrow$ 
    & LPIPS$\downarrow$ \\
    \midrule
    
    \multirow{2}{*}{Super-Resolution}
    & w/o & 46.42 & 23.30 & 0.6462 & 0.3396 \\
    & w/ & \textbf{46.10} & \textbf{23.35} & \textbf{0.6476} & \textbf{0.3373} \\
    \midrule
    
    \multirow{2}{*}{Denoising}
    & w/o & 45.70 & 25.50 & 0.7215 & 0.3025 \\
    & w/ & \textbf{45.29} & \textbf{26.18} & \textbf{0.7437} & \textbf{0.2857} \\
    \midrule
    
    \multirow{2}{*}{Inpainting}
    & w/o & 47.04 & 22.33 & 0.6318 & 0.3611 \\
    & w/ & \textbf{45.64} & \textbf{24.82} & \textbf{0.7150} & \textbf{0.2990} \\
    \midrule
    
    \multirow{2}{*}{Colorization}
    & w/o & 47.21 & 21.88 & 0.7220 & 0.3845 \\
    & w/ & \textbf{45.97} & \textbf{22.19} & \textbf{0.7465} & \textbf{0.3654} \\
    \bottomrule
\end{tabular}
\end{table*}

\medskip
\noindent
\textbf{Perceptual Network.}
We compare the effect of using external and internal perceptual networks for perceptual supervision, as summarized in \cref{tab:ablation_percep_network}. Note that external network corresponds to our reimplementation of E-LatentLPIPS, and the internal network corresponds to using intermediate and final decoder features. Using the external perceptual network improves distortion metrics such as PSNR and SSIM. However, applying the proposed gradient alignment with the following external network does not lead to further improvements and instead yields the model to deviate from the optimal trajectory. This suggests that perceptual supervision derived from the VGG network trained on an external dataset may not be suitable for restoration dynamics. 

In contrast, defining the perceptual objective based on internal decoder features leads to improved perceptual alignment with the restoration objective. In particular, incorporating the proposed conflict-free gradient alignment further reduces FID from 46.01 to 45.50, yielding the best perceptual quality among all configurations. The following results suggest that internal perceptual representations are better aligned with the latent restoration dynamics, and that the proposed gradient alignment strategy further stabilizes the interaction between reconstruction and perceptual objectives. Hence, we adopt the internal perceptual network with conflict-free gradient alignment as our final configuration.

\begin{table}[t]
\centering
\footnotesize
\setlength{\tabcolsep}{4pt}
\renewcommand{\arraystretch}{1.12}
\caption{\textbf{Ablation on Perceptual Network.} Comparison between external perceptual supervision (E-LatentLPIPS) and internal decoder feature supervision for super-resolution task. Conflict-free gradient alignment further improves perceptual quality when applied to the internal perceptual network, achieving the best FID.}
\label{tab:ablation_percep_network}

\begin{tabular}{l cccc}
    \toprule
    Perceptual Network
    & FID$\downarrow$ 
    & PSNR$\uparrow$
    & SSIM$\uparrow$ 
    & LPIPS$\downarrow$ \\
    \midrule
    
    External
    & \textbf{46.03} & \textbf{23.58} & \textbf{0.6604} & \textbf{0.3223} \\
    $+$ Conflict-Free Gradient Alignment & 46.15 & 23.41 & 0.6514 & 0.3299 \\
    \midrule
    
    Internal
    & 46.01 & 23.31 & \textbf{0.6515} & \textbf{0.3310} \\
    $+$ Conflict-Free Gradient Alignment & \textbf{45.50} & \textbf{23.38} & 0.6512 & 0.3328 \\
    \bottomrule
\end{tabular}
\end{table}

\medskip
\noindent
\textbf{Perceptual vs. Structural Gradient Projection.}
We further compare two conflict-resolution strategies: projecting the perceptual gradient ($g_{\mathrm{LCPL}} \rightarrow \tilde g_{\mathrm{LCPL}}$) and projecting the structural gradient ($g_{\mathrm{LCFM}} \rightarrow \tilde g_{\mathrm{LCFM}}$). As shown in \cref{tab:ablation_gradient_projection}, projecting the structural gradient generally yields better FID across all restoration tasks. These results support our design choice of treating the perceptual objective as a steering signal and resolving optimization conflicts through structural gradient projection. We hypothesize that perceptual supervision primarily guides the transport trajectory toward perceptually realistic regions of the target manifold, whereas the structural objective provides reconstruction-oriented updates. In addition, we observe that the perceptual gradient exhibits a larger magnitude than the structural gradient, which may further contribute to the improved optimization stability of structural-gradient projection.

\begin{table}[htbp]
\centering
\footnotesize
\setlength{\tabcolsep}{4pt}
\renewcommand{\arraystretch}{1.12}
\caption{\textbf{Ablation on Gradient Projection.} Quantitative comparison between projecting the perceptual gradient and projecting the structural gradient during conflict-free gradient alignment. Projecting the structural gradient $g_{\mathrm{LCFM}} \rightarrow \tilde{g}_{\mathrm{LCFM}}$ generally yields better model performance in FID across restoration tasks.}
\label{tab:ablation_gradient_projection}

\begin{tabular}{lccc}
    \toprule
    Task & Gradient Projection
    & FID$\downarrow$ 
    & LPIPS$\downarrow$ \\
    \midrule
    
    \multirow{2}{*}{Super-Resolution}
    & $g_{\mathrm{LCPL}} \rightarrow \tilde{g}_{\mathrm{LCPL}}$ & 45.66 & \textbf{0.3317} \\
    & $g_{\mathrm{LCFM}} \rightarrow \tilde{g}_{\mathrm{LCFM}}$ & \textbf{45.50} & 0.3328 \\
    \midrule
    
    \multirow{2}{*}{Denoising}
    & $g_{\mathrm{LCPL}} \rightarrow \tilde{g}_{\mathrm{LCPL}}$ & 45.42 & 0.2800 \\
    & $g_{\mathrm{LCFM}} \rightarrow \tilde{g}_{\mathrm{LCFM}}$ & 45.42 & 0.2800 \\
    \midrule
    
    \multirow{2}{*}{Inpainting}
    & $g_{\mathrm{LCPL}} \rightarrow \tilde{g}_{\mathrm{LCPL}}$ & 45.53 & \textbf{0.2935} \\
    & $g_{\mathrm{LCFM}} \rightarrow \tilde{g}_{\mathrm{LCFM}}$ & \textbf{45.50} & 0.2936 \\
    \midrule
    
    \multirow{2}{*}{Colorization}
    & $g_{\mathrm{LCPL}} \rightarrow \tilde{g}_{\mathrm{LCPL}}$ & 45.21 & 0.3599 \\
    & $g_{\mathrm{LCFM}} \rightarrow \tilde{g}_{\mathrm{LCFM}}$ & 45.21 & \textbf{0.3596} \\
    \bottomrule
\end{tabular}
\end{table}

\subsection{Further Analysis}
\begin{figure*}[htbp]
    \centering
    \includegraphics[width=0.9\linewidth]{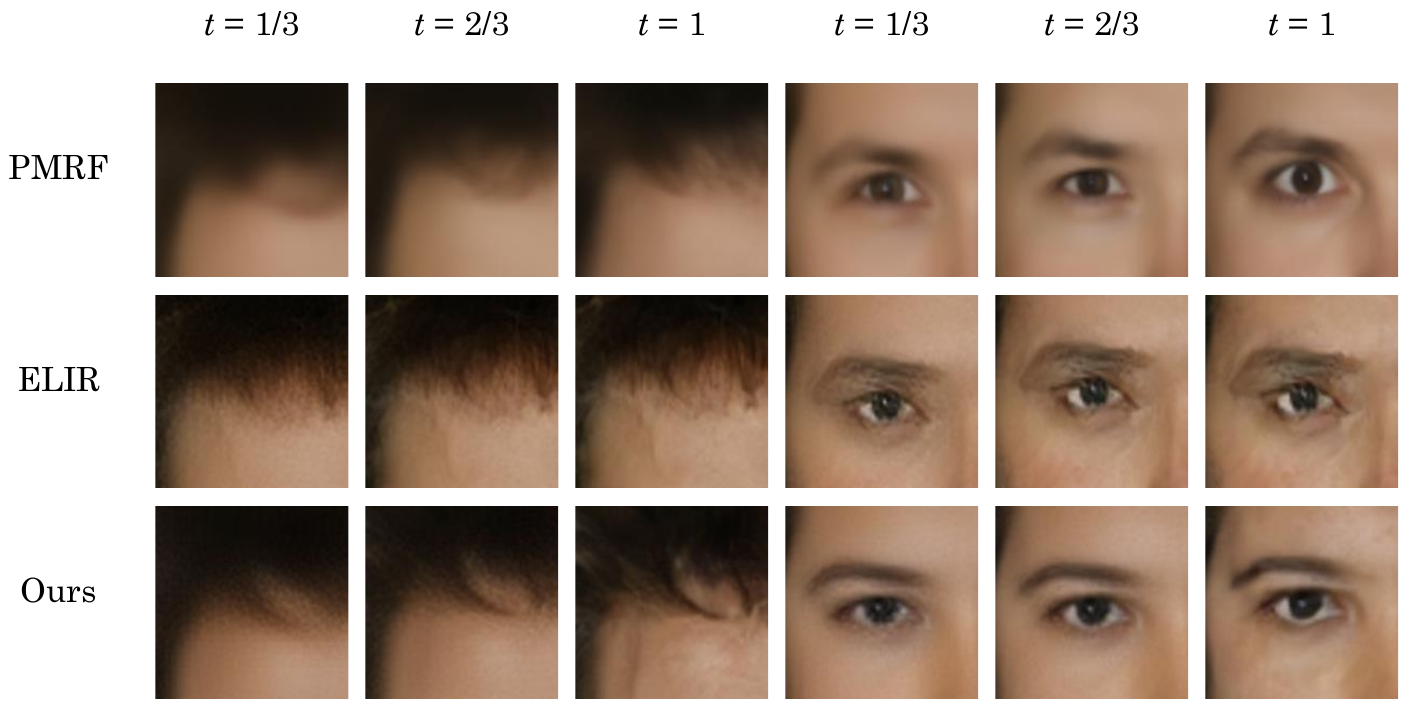}
    \caption{\textbf{Comparison in trajectory of PCFlow with its baselines.} PCFlow exhibits a stable coarse-to-fine refinement trajectory, where global structure is initially recovered in early steps and fine details are progressively introduced. In contrast, PMRF produces inconsistent intermediate states, while ELIR introduces perceptual details prematurely, often leading to structurally inconsistent textures.}
    \label{fig:analysis}
\end{figure*}

\cref{fig:analysis} compares the transport trajectories of PCFlow compared with its baselines, ELIR \cite{Cohen2025BMVC} and PMRF \cite{ohayon2025posteriormean}. PMRF exhibits unstable structural progression across timesteps. Early predictions remain overly smooth and high-frequency details appear abruptly, resulting in inconsistent intermediate states. ELIR attempts to introduce perceptual details from the beginning of the trajectory. However, these details are structurally inconsistent, suggesting that the model prioritizes texture synthesis before sufficiently recovering the underlying structure. In contrast, PCFlow follows a coarse-to-fine restoration trajectory. PCFlow primarily recovers the global facial structure in early steps, whereas high-frequency details such as hair texture and eye boundaries are progressively refined in later timesteps. 

\subsection{Further Qualitative Results}

\begin{figure}[htbp]
    \centering
    \includegraphics[width=0.9\linewidth]{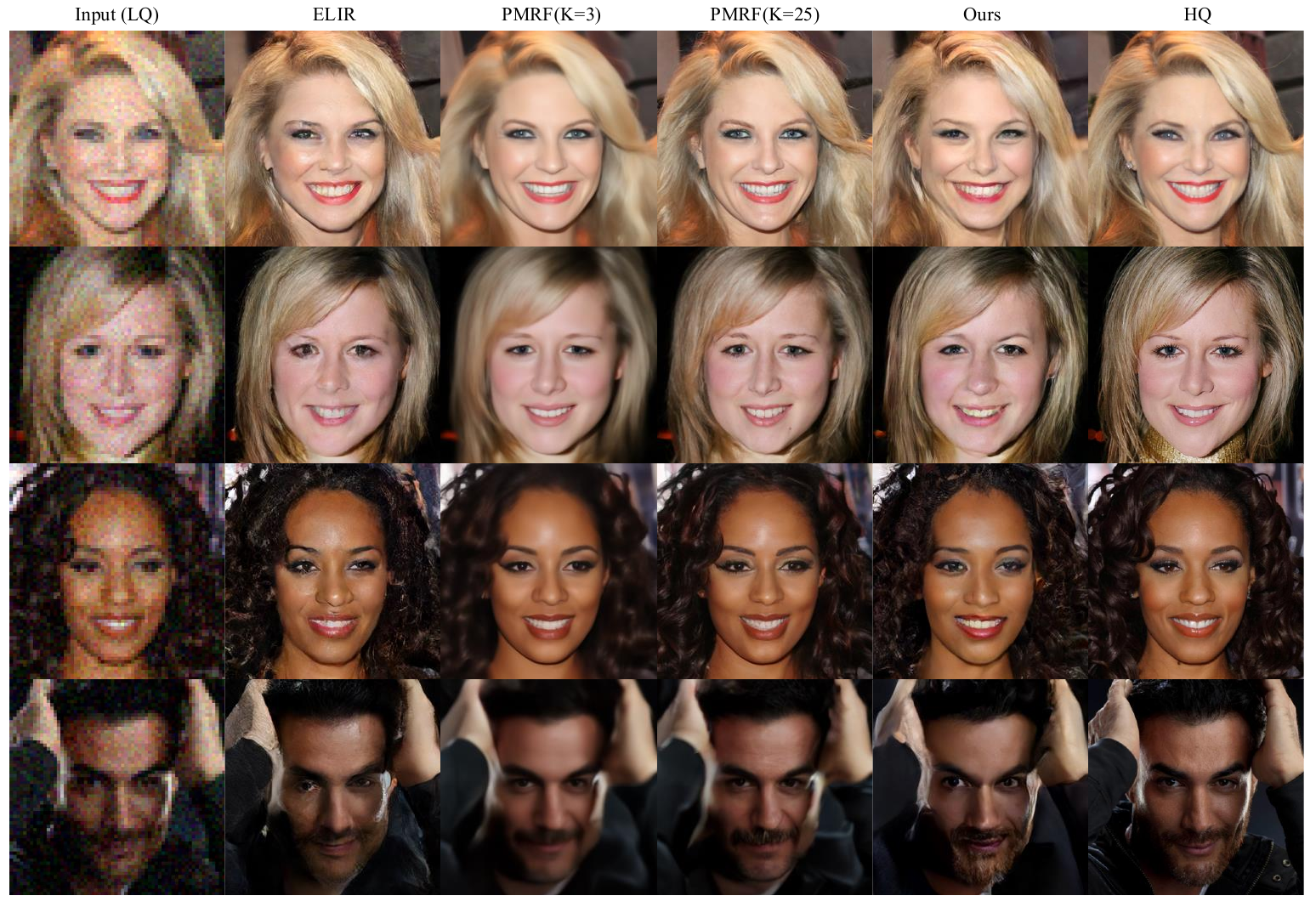}
    \caption{\textbf{Additional Results of PCFlow on super-resolution task.} From left to right: Input(LQ), ELIR, PMRF($K = 3$), PMRF($K = 25$), PCFlow(Ours), and ground truth image(HQ).}
    \label{fig:additional_fig_sr}
\end{figure}

\begin{figure}[htbp]
    \centering
    \includegraphics[width=0.9\linewidth]{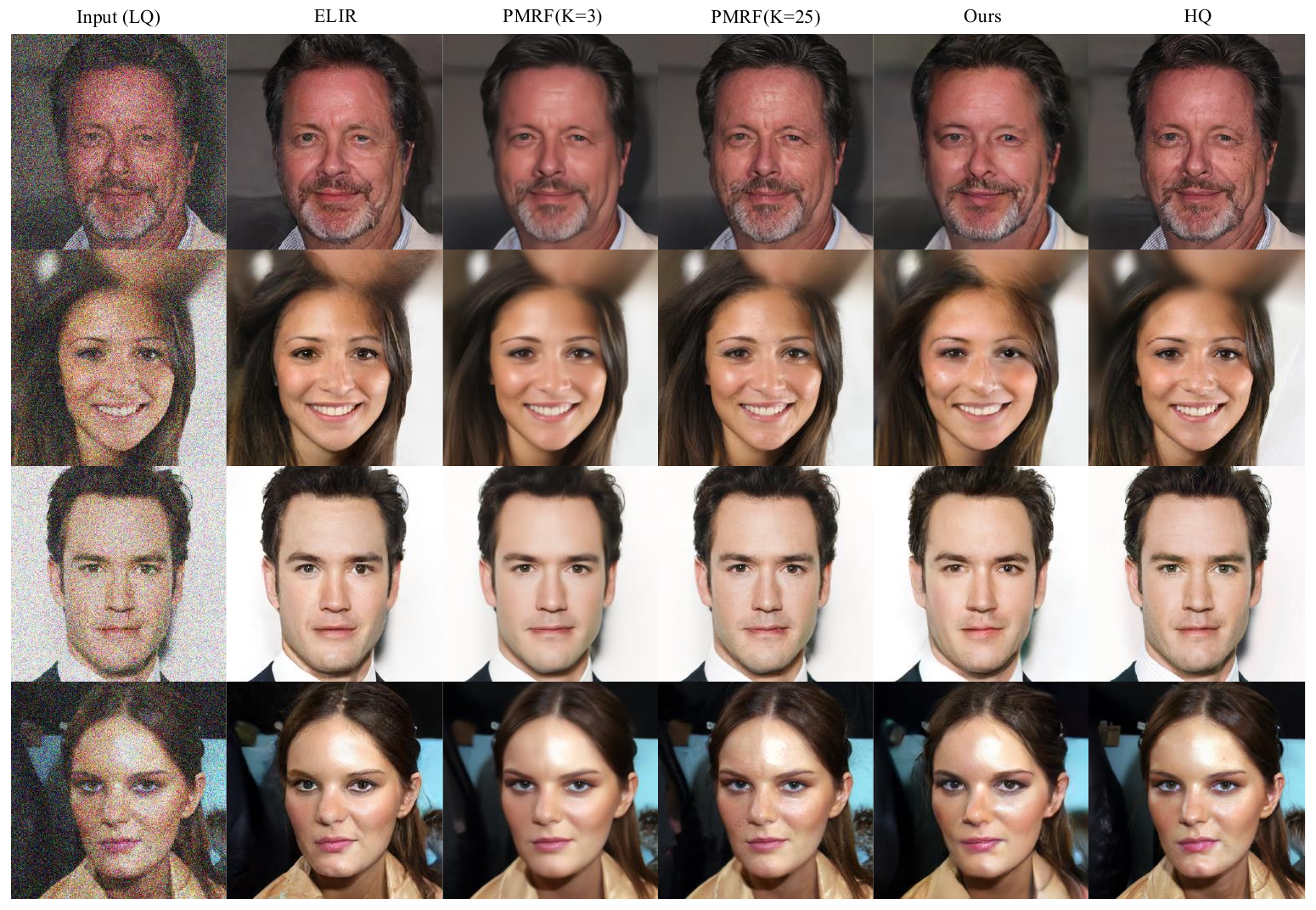}
    \caption{\textbf{Additional Results of PCFlow on denoising task.} From left to right: Input(LQ), ELIR, PMRF($K = 3$), PMRF($K = 25$), PCFlow(Ours), and ground truth image(HQ).}
    \label{fig:additional_fig_denoising}
\end{figure}

\begin{figure}[htbp]
    \centering
    \includegraphics[width=0.9\linewidth]{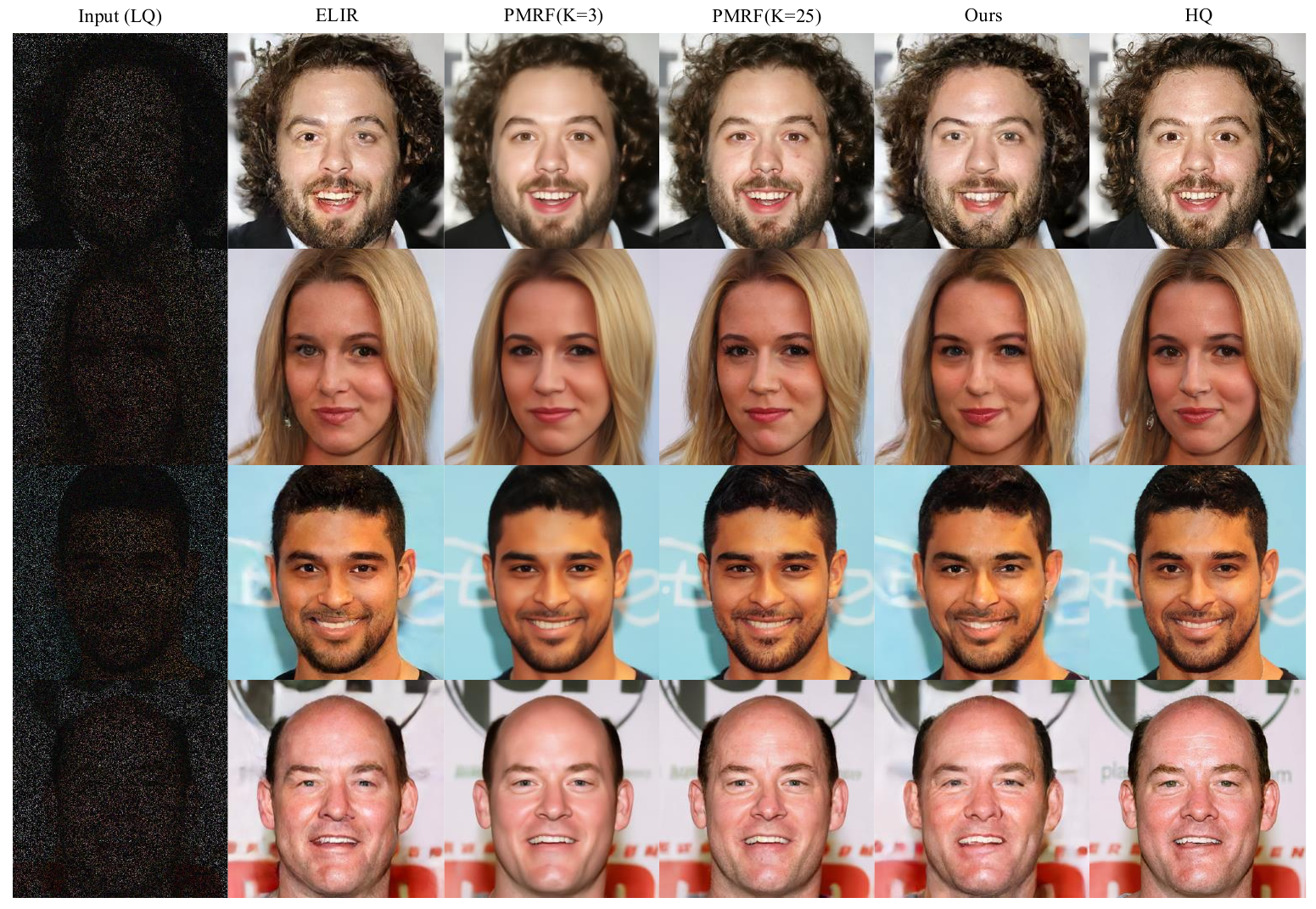}
    \caption{\textbf{Additional Results of PCFlow on inpainting task.} From left to right: Input(LQ), ELIR, PMRF($K = 3$), PMRF($K = 25$), PCFlow(Ours), and ground truth image(HQ).}
    \label{fig:additional_fig_inpainting}
\end{figure}

\begin{figure}[htbp]
    \centering
    \includegraphics[width=0.9\linewidth]{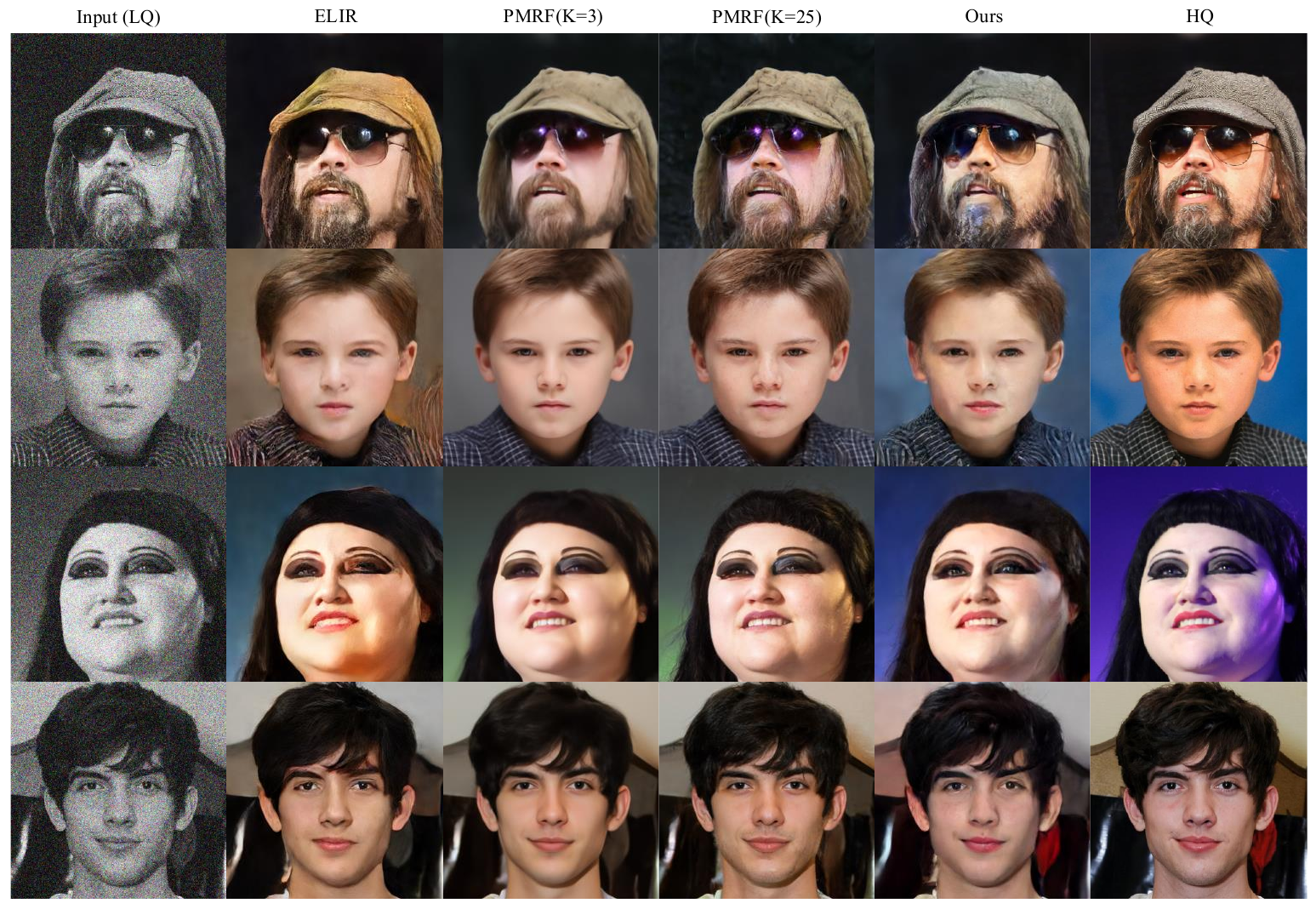}
    \caption{\textbf{Additional Results of PCFlow on colorization task.} From left to right: Input(LQ), ELIR, PMRF($K = 3$), PMRF($K = 25$), PCFlow(Ours), and ground truth image(HQ).}
    \label{fig:additional_fig_colorization}
\end{figure}



\end{document}